\documentclass[a4paper]{article}
\pdfoutput=1 
\usepackage{INTERSPEECH2020}
\newcommand\ft{\textsc{FT~Speech}}
\newcommand\danpass{DanPASS}
\newcommand\sprak{SBRead}
\newcommand\sbdictate{SBDictate}

\newcommand\SBrecipe{\texttt{sprakbanken}}

\title{\textsc{FT Speech}: Danish Parliament Speech Corpus}

\name{Andreas Kirkedal$^{1,2*}$\thanks{$^*$ Authors contributed equally.}, Marija Stepanovi\'{c}$^{1*}$\footnotemark[1], Barbara Plank$^1$}

\address{
  $^1$ IT University of Copenhagen (ITU), Copenhagen, Denmark\\
  $^2$ Interactions LLC, USA}
\email{aski@itu.dk,  maste@itu.dk,  bapl@itu.dk}

\begin{document}
\maketitle

\begin{abstract}
 This paper introduces \ft, a new speech corpus created from the recorded meetings of the Danish Parliament, otherwise known as the \textit{Folketing} (FT). 
 The corpus contains over 1,800 hours of transcribed speech by a total of 434 speakers. It is significantly larger in duration, vocabulary, and amount of spontaneous speech than the existing public speech corpora for Danish, which are largely limited to read-aloud and dictation data.
 We outline design considerations, including the preprocessing methods and the alignment procedure. To evaluate the quality of the corpus, we train automatic speech recognition systems (ASR) on the new resource and compare them to the systems trained on the Danish part of Språkbanken, the largest public ASR corpus for Danish to date. Our baseline results show that we achieve a 14.01 WER on the new corpus. A combination of \ft{} with in-domain language data provides comparable results to models trained specifically on Språkbanken, showing that \ft{} transfers well to this data set. Interestingly, our results demonstrate that the opposite is not the case. This shows that \ft{} provides a valuable resource for promoting research on Danish ASR with more spontaneous speech. 
\end{abstract}


\section{Introduction}

The main obstacle to effective automatic speech recognition is the lack of training material, especially for less-resourced languages such as Danish. Fortunately, the recent proliferation of available online content and open-source software has facilitated the collection of such data, lowering the threshold for wider-spread open ASR technology and research. In this regard, recordings of parliamentary sessions and their official reports are an invaluable resource. They are well-curated, ever growing, and openly accessible. 
Nevertheless, parliamentary speeches and their transcripts are not readily usable for ASR due to issues such as inaccurate timestamps, non-verbatim transcripts, and overly long utterances.

In this paper, we describe the development of a new ASR resource for Danish, \ft{}. It represents the biggest speech corpus for Danish spanning nine years of source material (2010--2019) and advancing Danish from a medium-resource to a high-resource language with respect to open-access speech data~\cite{lacunae}. We evaluate baselines for the new corpus and compare them to the ones trained on existing resources. 
Since parliamentary recordings are released on an ongoing basis, we plan to update the corpus as more source data becomes available.

To ensure replicability, we will release the code required to reproduce the corpus creation and evaluation from scratch. At the same time, to ensure accessibility, we will also provide the resulting timestamps and transcripts that can be used to extract the corpus utterances directly from the meeting recordings. All materials we provide are freely available for research purposes only. The data, license, and terms of use can be found on \url{ftspeech.dk}.\footnote{We thank the Danish Parliament for making their data available.}

\section{Related Work}

Research into ASR for Danish has been rather limited owing to the scarcity of publicly available speech corpora~\cite{lacunae}. At present, there are only two  public Danish speech corpora: Danish Språkbanken and \danpass{}, which were developed under different research questions and objectives.

The more comprehensive of the two, Danish Språkbanken, is included in Språkbanken (Norwegian Language Bank), a collection of open-access and open-source language resources for Norwegian, Swedish, and Danish compiled by Nordisk Språkteknologi (NST). It contains two subsets designed specifically for the development of ASR systems: NST Danish ASR Database (\sprak{})~\cite{sprakbanken} and NST Danish Dictation (\sbdictate{})~\cite{sprakbanken_dictate}. 

\sprak{} is the only public data set suitable for training ASR systems. It contains around 390 hours of phonetically balanced read-aloud speech by a total of 616 speakers, as well as general meta-data about the speakers. A standardized version of this data set was released with a recipe to train ASR systems in the Kaldi repository (\SBrecipe{})~\cite{ASKphd2016}. On the other hand, \sbdictate{} is a smaller data set with roughly 54 hours of speech by 151 speakers aimed at acoustic modeling of automatic dictation. However, despite their size and speaker variety, both of these data sets are limited by their highly contrived nature. Namely, the utterances in these data sets constitute read-aloud sentences or phrases such as personal names, place names, acronyms, numerals, spelled out letters, etc.

\danpass{} is a phonetically annotated speech corpus primarily intended for acoustic and auditory phonetic analyses~\cite{gronnum2006danpass}. It contains about 9 hours of speech by 27 speakers recorded in a studio using professional recording equipment. Although this corpus may offer a theoretical basis for the development of speech technologies, it is not particularly suitable for ASR due to its small size and artificial setting. 
 
Currently, the Kaldi recipe \SBrecipe{} represents the state-of-the-art on the \sprak{} test set and \danpass{}~\cite{kirkedal-2018-acoustic}. However, for reasons outlined above, the perfomance of these models degrades sharply when they are confronted with large-vocabulary spontaneous speech, as we will show in Section \ref{sec:asr-experiments}.

In order to compile a corpus of utterances more akin to rapid spontaneous speech, we follow the recent trend of converting open parliamentary data into ASR speech corpora. This has been accomplished for languages such as Icelandic~\cite{helgadottir2017building}, Finnish~\cite{mansikkaniemi2017automatic}, and Bulgarian~\cite{bulgarian_asrcorpus}. 
In addition, a multilingual speech corpus has been constructed from the debates held in the European Parliament~\cite{europarlST}. Meanwhile, the official reports of Folketing meetings have already been used to create a text corpus released within CLARIN~\cite{DanishParliamentCorpus} and proved invaluable for multiple research disciplines~\cite{HaltrupFTGenderStudy,pedersen-etal-2016-semdax, hansen2019towards}. 

In constructing \ft{}, we follow a procedure similar to the one used to create LibriSpeech~\cite{LibriSpeech}. Other influential work on automatic alignment methods in the creation or correction of speech corpora includes \cite{alignment_Hazen2006, alignment_haubold2007, alignment_limited_resources2014, Montreal_aligner2017}.

\section{Corpus Preparation and Alignment}

This section presents the corpus preparation and alignment procedures, including the description of the raw source data, audio and text preprocessing, lexicon creation, and alignment.

\subsection{Source Data Description}\label{sec:datadesc}

\ft{} was created from the recordings of Danish parliamentary sessions and their annotated reports, which are freely available on the Folketing's official website: \texttt{ft.dk}. The audio recordings are available in two formats: \texttt{MP3} (audio only) and \texttt{AAC} (as part of the audio and video stream container \texttt{MP4}).

The sessions used to create the corpus include 1,003 meetings of the Parliament recorded in the period from October 5, 2010 (first video broadcast) until December 20, 2019 (last meeting in 2019). This amounted to about 4,960 hours of recorded audio material featuring 447 different speakers. 

The reports of the parliamentary meetings are transcribed and published by the Office of the Folketing Hansard. Each report contains a comprehensive account of all parliamentary activities in the course of one meeting, including near-verbatim transcripts of the speeches by Members of Parliament (MPs) accompanied by their corresponding metadata, such as the speaker's name, role, and political affiliation, as well as the approximate start and end timestamps of the speech. 

The reports are released online as \texttt{XML} and \texttt{PDF} documents.
Initially, only a preliminary version is released while the report is still subject to revision. From this point forward, it can take up to several months until the final version is published. During this period, the reports, and, in particular, the speech transcripts may undergo a number of modifications to ensure adherence to the formal guidelines established by the Presidium of the Danish Parliament. Therefore, the speeches are not transcribed strictly verbatim, but are instead adapted into standard written form by omitting speech disfluencies, correcting factual errors and slips of the tongue, and adding context to ensure the transcripts reflect the intentions of the speaker clearly and accurately \cite{fttidende}.

In addition to prescribing documentation guidelines, the Presidium also enforces observance of parliamentary etiquette, which mandates decorum and the use of formal and respectful language in the Parliament. Some of the official rules state that the MPs must be addressed as either \textit{Mr.} or \textit{Ms.} followed by their full name, while the Ministers must be addressed with their minister titles. Furthermore, the MPs may not interject, applaud, express disapproval, or otherwise make noise during speeches and debates\footnote{A parliamentary debate is a sequence of monologues on the same topic.}~\cite{HaltrupFTGenderStudy}. This makes the FT meetings well-suited for the extraction of speaker-annotated monologues used in ASR research and development. However, the recordings still occasionally contain an audible level of spoken background noise. 

The speakers come from different administrative regions of Denmark, as well as the two autonomous territories within the Kingdom of Denmark: Greenland and the Faroe Islands. Although some of the speakers may be native speakers of other local languages or dialects, the official language of the Parliament is Danish. In particular, since the linguistic register is strictly formal, while the topics discussed are primarily concerned with social, political, economic, and legal matters, the idiolects used in the Parliament converge on Standard Danish. The manner of delivery ranges from read or rehearsed to spontaneous speech.

The main challenges of converting this kind of raw data into a corpus suitable for ASR stem from: 
1) the inaccuracy of the timestamps indicating the beginning and end of speeches in the reports by up to 30 seconds, 
2) discrepancy between the written transcripts and the actual speeches,
3) presence of background noise in the audio data, and
4) use of lossy compression formats (\texttt{MP3} and \texttt{AAC}) to encode the audio data.

\subsection{Audio and Text Preprocessing}\label{preprocessing}

First, we downloaded the audio recordings of all FT meetings available on the official website up to and including December 2019.\footnote{URL to the  video and audio recordings: \\ \texttt{ft.dk/da/dokumenter/dokumentlister/referater}}
All the recordings were in the \texttt{MP3} format with a bitrate of 128 kbit/s. Their duration ranged from 5 minutes to 16 hours (mean $\approx 5$ h, SD $\approx 3$ h).

We began the audio processing by extracting the left channel stream from the stereo recordings. This was an arbitrary decision since the two channels were identical. The mono recordings were left unchanged at this stage. Next, we converted the selected single-channel \texttt{MP3} recordings to \texttt{WAV} using a 16-bit linear PCM sample encoding (PCM\_S16LE) sampled at 16~kHz.
Finally, to extract speeches by single speakers, we segmented the obtained \texttt{WAV} files according to the timestamps and speaker names provided in the annotated meeting reports in the \texttt{XML} format.\footnote{URL to \texttt{XML} transcripts of the proceedings: \\ \texttt{ftp://oda.ft.dk/ODAXML/Referat/samling}} To ensure the speaker names in the annotations referred to unique individuals, we cross-checked them with the biographies of past and present MPs available on the official website.\footnote{URLs to the biographies of FT MPs in Danish and English: \\ \texttt{https://www.ft.dk/da/medlemmer}\\ \texttt{https://www.thedanishparliament.dk/en/members}}
This procedure resulted in 414K speeches whose duration ranged from less than 1 second to 15 minutes (mean $\approx 40.1$ s, SD $\approx 63.68$ s). However, most speeches did not perfectly align with their corresponding transcripts due to the inaccuracy of the timestamps in the \texttt{XML} annotations, 
which is one of the issues we try to overcome with the alignment procedure outlined in Section~\ref{sec:alignment}

As stated earlier, the textual transcripts of the speeches were extracted from the \texttt{XML} documents containing the reports of the FT meetings. Their preprocessing involved expanding all common abbreviations, numbers, dates, and symbols, as well as removing all punctuation, capitalization, and unspoken parenthetical remarks and references.

\begin{table*}[ht!]
  \caption{Corpus partitions and their size in hours, total number of utterances, tokens, types, OOV tokens, and speakers. The speaker counts by gender are marked as \textbf{F} (female) and \textbf{M} (male). OOV tokens comprise all tokens whose types are not part of our ASR lexicon.}
  \label{tab:corpus_subsets}
  \centering
  \begin{tabular}{lrrrrrr}
    \toprule
    \textbf{Subset}  &  \textbf{Hours}  & \textbf{Utterances} &  \textbf{Tokens} & \textbf{Types} & \textbf{OOV tokens} & \textbf{Speakers (F+M)} \\
    \toprule
    train      & 1,816.29  & 995,677    & 18,865,071 & 147,326 & 0 & 374 (146+228)  \\
    \midrule
    dev-balanced        & 5.03     & 2,601      & 51,497     & 6,888   & 2,846 & \multirow{2}{*}{20 (10+10)}      \\
    dev-other  & 14.96    & 7,595      & 152,225    & 12,701  & 8,248    \\
    \midrule
    test-balanced       & 10.05    & 5,534      & 103,439    & 10,050  & 5,871 & \multirow{2}{*}{40 (20+20)}      \\
    test-other & 10.88    & 5,837      & 111,818    & 10,491  & 6,145   \\
    \midrule
    total     & 1,857.21   & 1,017,244  & 19,284,050  & 149,239 & 23,110 & 434 (176+258)  \\
    \bottomrule
  \end{tabular}
\end{table*}

\subsection{Alignment Lexicon}

The lexicon used for alignment was produced by concatenating the vocabulary created from the preprocessed transcripts of FT speeches with the \SBrecipe{} lexicon containing all words from the \sprak{} train set. This yielded around 233K unique words (types). Their pronunciations were generated using eSpeak NG,\footnote{\texttt{https://github.com/espeak-ng/espeak-ng}} a multilingual rule-based grapheme-to-phoneme converter and speech synthesizer. We stripped these pronunciations off all stress, vowel length, and stød markers. We also made the pronunciations of foreign words consistent with the Danish phonetic alphabet in eSpeak, and manually added the unstressed forms of common function words.

\subsection{Alignment Model}\label{sec:alignment}

Because the FT meeting reports are not transcribed verbatim, we expect a large proportion of words in the aligned utterances to be misaligned. For instance, if a speaker mistakenly stated, \emph{My uh party is against tax- taxation of the air used to to create soft ice}, but the party were, in fact, in favor of such taxation, the transcript would be edited by changing \emph{against} to \emph{for} and removing filler words (\emph{uh}), restarts (\emph{tax-}), and repetitions (\emph{to}). In this example, if \emph{against} were correctly recognized by an ASR system, the word would be misaligned because it did not match the transcript. For this reason, we need to extract verbatim transcriptions while allowing for word repetitions, restarts, and filler words that occur frequently in spontaneous speech. In our example, we would extract two segments: \emph{My party is} and \emph{taxation of the air used to create soft ice}.

First, we create a word alignment with a procedure similar to the one used in LibriSpeech.\footnote{Implemented in the script
\texttt{clean\_and\_segment\_data.sh}
in the Kaldi repository} To compare ASR hypotheses to transcripts, we decode the timestamp-segmented FT data with a speaker-independent GMM AM trained with boosted MMI on training data from \SBrecipe{}.\footnote{Training scripts will be made available on \url{ftspeech.dk}.} We use standard MFCC features and a GMM AM because we expect better performance on data from a mismatched domain than with a DNN AM. The LMs used to generate the word alignment are trained on groups of utterances and the most frequent words in the FT data. We want to bias the LMs to the training utterances to find a trade-off between accurate decoding of the verbatim segments and robustness to the text mismatch. 

Next, we segment the utterances again and keep segments that start with a correctly recognized word and end with a misrecognized word (\emph{against} in our example).
If the segmentation algorithm classifies a misrecognized word as a word repetition such as \emph{to}, the word is inserted into the reference and we do not split the segment. If segments contain silences longer than 0.3 seconds, they are split again.
All ends of segments are extended in the audio and transcript to include the next word and to reduce edge effects in feature extraction from the core segment. The transcript is padded with a silence token such as \texttt{<UNK>} 
unless the segment is at the utterance boundaries. 

We restrict the utterances we include in the corpus to utterances with a duration ranging from 2 to 60 seconds to ensure they can be used to train AMs. As a result, we discard 487,938 utterances of which 1,320 were longer than 1 minute.

\section{Corpus Description and Organization}

\noindent Following the alignment, segmentation, and elimination of overly long or short utterances, the finished corpus, termed \ft{}, contains 1,017,244 utterances with a total duration of 1,857 hours produced by 434 speakers.

We partition this corpus into a training, development, and test set with no speaker overlap. To create the development and test set, we select the same number of male and female speakers with at least 150 utterances and 900 seconds per speaker, while trying to minimize training data loss. 

Since the total duration per speaker varies significantly, it was impossible to create a speaker-balanced development or test set. Therefore, we decided to further partition both of these sets into two subsets: \textit{balanced} and \textit{other}. The balanced portions (\textit{dev-balanced} and \textit{test-balanced}) contain approximately equal amounts of speech per speaker. They were created by choosing a random sample of utterances from each speaker such that the total duration per speaker was 900 seconds and that the difference in the number of utterances per speaker was kept low. The other portions (\textit{dev-other} and \textit{test-other}) consist of the remaining utterances by the same speakers which had to be removed from the training data to avoid speaker overlap.
Detailed statistics of the corpus and each of the partitions are shown in Table~\ref{tab:corpus_subsets}.

\section{Speech Recognition Experiments} \label{sec:asr-experiments}

This section describes the ASR experiments conducted for the purpose of evaluating the new resource, \ft{}.
We build two acoustic models: one trained on \ft{} train (FT AM) and the other on \sprak{} train data (SB AM), as well as two language models: one trained on FT text data (FT LMs) and the other on \sprak{} training transcripts (SB LMs), and subsequently evaluate their in-domain and out-of-domain combinations on three different test sets. Since \sprak{} is an established ASR corpus, we use the acoustic and language models trained on it as a reference point in our performance evaluation.

\subsection{Acoustic Models}\label{sec:acoustic_model}

We follow the \SBrecipe{} recipe\footnote{The recipe can be found here: \url{https://github.com/kaldi-asr/kaldi/blob/master/egs/sprakbanken/s5/run.sh}} to train monophone and triphone segmentation GMM AMs~\cite{gmmAmOriginal} from scratch on \ft{} and generate an alignment to train an iVector model~\cite{ivector} for speaker adaptation of a Time-Delay Neural Network (TDNN)~\cite{peddinti2015tdnn} AM. The only modification compared to the \SBrecipe{} recipe is that we do not perform data augmentation with speed-perturbation on \ft{} because the size of the training data is larger than the size of \sprak{} augmented with speed-perturbation. 
For \sprak{}, we use the training, development, and test split introduced in~\cite{kirkedal-2018-acoustic}. 

We train so-called \emph{chain} TDNN AMs with the LF-MMI objective on \ft{} and \sprak{}. LF-MMI is a sequence discriminative training criterion that maximizes the log probability of the correct phone sequence~~\cite{povey2016purely}.
We train the AMs for 4 epochs on minibatches of 128 chunks where each chunk contains 150 frames. The frames are 40-dimensional MFCC features.
The feature frames are subsampled so we only train on every third frame, but we create different versions of the training data by shifting the frames by 1 and 2 frames to create 3 versions. The effect is that every training epoch corresponds to 3 epochs. 
We use HMMs with a single state rather than the classic 3-state topology because of the low frame rate. 

The first layer of the TDNN stacks 3 frames and a 100-dimensional iVector and projects the supervector to a 450-dimensional vector with an affine transform. The remaining layers consists of an affine transform, ReLU activation and a \emph{renorm} component which is a layernorm without the mean term. 
We use a learning rate that decays from 0.001 to 0.0001 during training and we clip parameters at a Frobenius norm of 2.0. 
All hyperparameters are copied from the \SBrecipe{} recipe and are identical for the two AMs. Note that there are several important differences to the AM in \cite{kirkedal-2018-acoustic} (see section 7.1).

\subsection{ASR Lexicon and Language Models}\label{sec:language_models}

To create the lexicon for the ASR experiments, we reuse the alignment lexicon but remove all types that appear only in the preprocessed transcripts of speeches by the speakers placed in the \ft{} development and test sets.\footnote{Note that neither the ASR nor the alignment lexicon contains any \sprak{} test or development data.} 
With SRILM~\cite{srilmStolcke02}, we estimated several 3-gram and 4-gram LMs with Witten-Bell or Kneser-Ney smoothing on both text from the FT meeting reports and on the \sprak{} transcripts. Ultimately, we choose the trigram models with Witten-Bell smoothing, which we will refer to as FT~LM and SB~LM, for the final evaluation, as they achieve the best performance on their corresponding development sets. Before training, the transcripts were segmented into sentences using the spaCy sentence segmenter for Danish \cite{spacy2}, and then preprocessed as described in Section \ref{preprocessing}.

\section{Performance Evaluation}\label{sec:performance}

We use the standard word error rate metric (WER) to evaluate the performance. Our evaluation spans all four combinations of the two AMs and the two LMs (the lexicon is constant in all cases).
We evaluate each of them on three test sets: \sprak{} test (introduced in~\cite{kirkedal-2018-acoustic} as SPTEST), \sbdictate{} test (included in the original \sbdictate{} data~\cite{sprakbanken_dictate}), and \ft{} test-balanced (Table~\ref{tab:corpus_subsets}). The WER results for all combinations of AMs, LMs, and test data are shown in Table \ref{tab:eval_results_v1}.

\begin{table}[th!]
  \caption{WER performance of all four AM+LM combinations on three different test sets.} 
  \label{tab:eval_results_v1}
  \centering
  \begin{tabular}{llrr}
    \toprule
    \textbf{AM}           &  \textbf{Test set}      & SB LM         & FT LM           \\
    \midrule
    \multirow{3}{*}{SB}   & \sprak{}                & \textbf{8.81} & 15.98           \\ 
                          & \sbdictate{}            & 14.46         & 19.77           \\ 
                          & \ft{}                   & 37.52         & 23.86           \\  
    \midrule
    \multirow{3}{*}{FT}   & \sprak{}                & 13.07         & 27.22           \\  
                          & \sbdictate{}            & 20.71         & 33.73           \\ 
                          & \ft{}                   & 24.25         & \textbf{14.01}  \\ 
    \bottomrule
  \end{tabular}
\end{table}

From Table \ref{tab:eval_results_v1}, we can see that for \ft{}, the best WER of 14.01\% is obtained with the in-corpus LM and AM.  As expected, in-domain AM and LM combinations perform best in all in-domain settings (boldface in Table~\ref{tab:eval_results_v1}).\footnote{Our system SB AM+LM achieves a new best result on \sprak{}.} However, going across domains remains a challenge.  

When the LM does not match the domain of the test set, WER rises by 5--14\% absolute, presumably due to significant lexical differences between \sprak{} and \ft{}. As stated previously, a large number of \sprak{} utterances consist solely of either proper nouns, numerals, spelled out names, or imperative sentences, while some also contain articulated punctuation symbols used for modeling automatic dictation. These kinds of utterances, most of which were devised to increase phonetic diversity and type-to-token ratio, do not occur in \ft{} nor general spontaneous speech. For these reasons, \sprak{} is a less challenging resource, reflected in the lower WER (8.81).

We see a gap in performance when we fix the test set and LM, but replace the AM. This decrease in performance occurs as a result of the acoustic differences between the \ft{} and \sprak{} utterances, especially, the differences in the speech genre, recording environment and equipment, and audio encodings. Namely, \sprak{} was recorded in a quiet office and encoded in a lossless format, whereas \ft{} was recorded in the FT meeting chamber and encoded into a lossy format. 

Most importantly, however, we observe that the combination of FT AM and SB LM evaluated on \sprak{} test achieves a WER of 13.07\%, which is comparable to the results previously published on this test set (13.08--13.38\% WER, presented in Table 7 in~\cite{kirkedal-2018-acoustic}).  On  \sbdictate{}, it achieves a WER of 20.71  with FT AM+SB LM, which is 6.25\% absolute WER off from in-domain data results. This means that the FT AM generalizes well to \sprak{} data, and it shows the benefit of the new corpus containing more spontaneous speech in a more realistic environment with disfluencies and background noise. Interestingly, the converse is not the case: the SB AM does not generalize to the \ft{} domain, resulting in the worst overall WER. This shows how poorly existing resources generalize, which further underlines the value of the proposed resource. While not strictly comparable, our WER results on FT are in similar ranges to related work on Icelandic, another Northern Germanic language, were a WER of 14.76\% was reported on parliamentary speeches~\cite{helgadottir2017building}.

\section{Conclusion}

This paper introduces \ft{}, a novel corpus for Danish ASR containing more than 1,800 hours of speech. It enriches the limited landscape of existing resources for Danish with a resource containing more spontaneous speech in challenging realistic conditions.
Our baseline results show that a combination of \ft{} with in-domain language data provides not only comparable results to prior work, but also a more challenging benchmark for future studies. As the source material expands naturally, we will update the corpus with new data.

\bibliographystyle{IEEEtran}
\bibliography{mybib}

\end{document}